\pdfoutput=1
\documentclass[11pt]{article}
\usepackage{graphicx}
\usepackage{multirow}
\usepackage{booktabs}
\usepackage{bm}
\usepackage{amssymb}
\usepackage{amsfonts}
\usepackage{amsmath}
\usepackage{algorithm}
\usepackage{algorithmic}

\usepackage{float} 

\usepackage{fancyhdr}
\usepackage{longtable}
\usepackage{pifont}
\usepackage{mathrsfs}
\usepackage[]{acl}

\usepackage{times}
\usepackage{latexsym}

\usepackage[T1]{fontenc}
\usepackage[utf8]{inputenc}

\usepackage{microtype}

\title{A Simple Temporal Information Matching Mechanism for Entity Alignment Between Temporal Knowledge Graphs}

\author{Li Cai$^{1,2}$, Xin Mao$^{1}$, Meirong Ma$^3$, Hao Yuan$^3$, Jianchao Zhu$^3$, Man Lan$^{1}$ \thanks{*Corresponding author}\\
$^1$School of Computer Science and Technology, East China Normal University \\
$^2$College of Computer Science and Technology, Guizhou University \\
$^3$Shanghai Transsion Co., Ltd \\
\texttt{caili2020stu@gmail.com, xmao@stu.ecnu.edu.cn} \\
\texttt{\{meirong.ma, hao.yuan, jianchao.zhu\}@transsion.com} \\
\texttt{mlan@cs.ecnu.edu.cn}}
  
\begin{document}
\maketitle
\begin{abstract}
Entity alignment (EA) aims to find entities in different knowledge graphs (KGs) that refer to the same object in the real world. Recent studies incorporate temporal information to augment the representations of KGs. The existing methods for EA between temporal KGs (TKGs) utilize a time-aware attention mechanism to incorporate relational and temporal information into entity embeddings. The approaches outperform the previous methods by using temporal information. However, we believe that it is not necessary to learn the embeddings of temporal information in KGs since most TKGs have uniform temporal representations. Therefore, we propose a simple graph neural network (GNN) model combined with a temporal information matching mechanism, which achieves better performance with less time and fewer parameters. Furthermore, since alignment seeds are difficult to label in real-world applications, we also propose a method to generate unsupervised alignment seeds via the temporal information of TKG. Extensive experiments on public datasets indicate that our supervised method significantly outperforms the previous methods and the unsupervised one has competitive performance.
\end{abstract}

\section{Introduction}

Knowledge graphs (KGs) describe facts of the real world in a structured form of triples $(h,r,t)$, where $h$ represents the head entity, $r$ represents the relation, $t$ represents the tail entity. KGs have drawn great research attention from the academia \cite{DBLP:conf/dsc/LinLJW21,DBLP:journals/tnn/JiPCMY22} and have been widely used to enhance downstream applications such as question answering \cite{DBLP:conf/acl/SaxenaTT20,DBLP:conf/wsdm/QiuWJZ20} and recommendation systems \cite{DBLP:conf/recsys/AnelliNSFM21,DBLP:conf/kdd/ZhouZBZWY20}. 

\begin{figure}[ht]
  \centering  
  \includegraphics[scale=0.34]{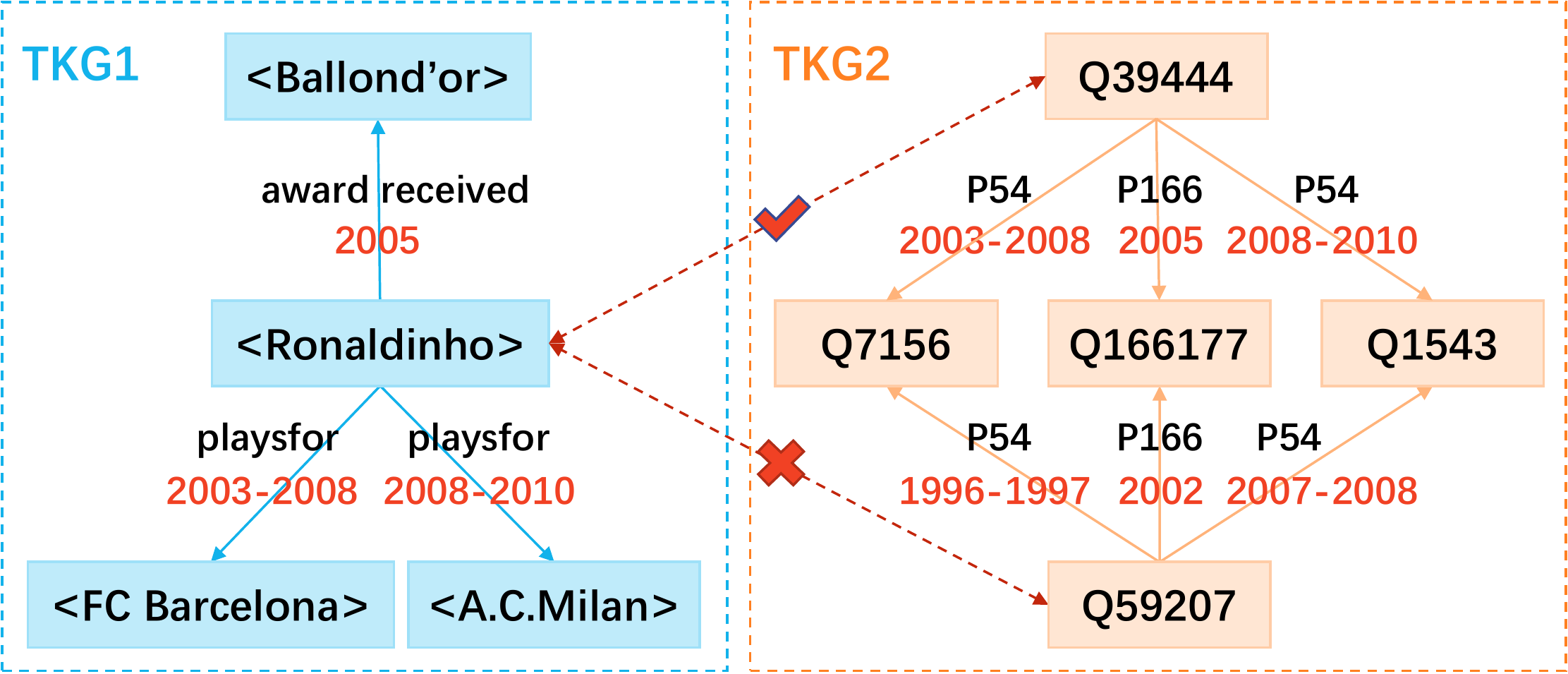}
  \caption{One sample of EA between TKGs. TKG1 is a sub-graph in YAGO. TKG2 is a sub-graph in WIKI, where the entity names corresponding to the items are as follows: \emph{Q39444 (Ronaldinho), Q7156 (FC Barcelona), Q166177 (Ballon d'Or), Q1543 (A.C.Milan), Q59207 (Ronaldo), P54 (member of sports team), P166(award received)}.} 
  \label{fig:sample}
\end{figure}

Recently, Wikidata \cite{DBLP:journals/cacm/VrandecicK14} and YOGO2 \cite{DBLP:journals/ai/HoffartSBW13} add temporal information to the relations to represent the KGs more accurately. In temporal knowledge graphs (TKGs), triples are extended to quadruples as $(h,r,t,\tau)$, where $\tau$ represents the timestamps. Figure~\ref{fig:sample} is an example of entity alignment (EA) between TKGs, where nodes represent entities, edges include relational and temporal information, and the quadruple \emph{(Ronaldinho, playsfor, A.C.Milan, 2008-2010)} represents one of the facts. 

EA seeks to find the same entities in the real world from different KGs to promote knowledge fusion. Many embedding-based methods perform effective EA \cite{DBLP:journals/pvldb/SunZHWCAL20,DBLP:conf/acl/MaoMYZWXWL22}, which encode entities in a continuous embedding space and align entities based on the learned embeddings. Previous EA methods ignore the temporal information despite their successes, a key factor indicating when a fact occurred. 

In order to utilize the time information in TKGs, TEA-GNN \cite{DBLP:conf/emnlp/XuS021} first proposes a time-aware GNN for EA between TKGs, which incorporates the temporal information via a time-aware attention mechanism to learn the embeddings of entities. TREA \cite{DBLP:conf/www/XuSX022} develops a temporal relational attention mechanism that introduces temporal embedding and relational embedding to entity embedding and achieves state-of-the-art (SOTA) performance.

In different TKGs, equivalent entity and relation pairs are usually disrupted and have different representation forms. As shown in Figure~\ref{fig:sample}, the entity of \emph{"Ronaldinho"} is stored in the form of \emph{<Ronaldinho>} in TKG1. But in TKG2, the same entity is stored as a pure id \emph{<Q39444>}. The relation of \emph{"plays for"} is stored in the form of \emph{<playsfor>} and <\emph{P54>} in the two TKGs respectively. Therefore, we need to map different representations into a unified vector space to infer whether they are similar. However, the temporal information is uniform across most TKGs. For example, the temporal information \emph{"2005"} is the same in the two TKGs, and refers to the same year. So we believe it's unnecessary to learn additional temporal embeddings. In addition, different entities can be distinguished by their temporal information even though they have similar structures and relations. Such as \emph{<Ronaldinho>} and \emph{Q59207 (Ronaldo)} in Figure~\ref{fig:sample}. In summary, we can directly use the temporal information for enhancing the prediction, rather than learning the temporal embeddings, which is redundant and time-consuming.

Based on this finding, we propose a \textbf{S}imple GNN model combined with a \textbf{T}emporal information matching mechanism for \textbf{E}ntity \textbf{A}lignment (STEA) between TKGs. Unlike the previous work, the model adopts a temporal information matching mechanism instead of learning temporal embeddings, which is effective and efficient. Our proposed model first takes a simple GNN to learn the embeddings of entities by using their structural and relational information, then compares the entities' temporal information, and finally aligns the entities by combining their embedding similarity and time similarity. Experimental results on four datasets show that our model significantly outperforms TREA \cite{DBLP:conf/www/XuSX022}. The improvement scores of Hits@1 are 3.4\%, 3.6\%, 2.2\%, 12.3\%, respectively.

Most existing methods of EA need alignment seeds to train the model and learn the embeddings of the entities. Finding the alignment seeds of different KGs in the real world is labor-intensive. Some unsupervised EA methods \cite{DBLP:conf/aaai/0001CRC21,DBLP:conf/ijcai/QiZCCXZZ21,DBLP:conf/emnlp/MaoWWL21} automatically generate alignment seeds via the image or name of entities. Inspired by these methods, we believe that uniform temporal information can also be used to generate alignment seeds. In this paper, we assume that entities with the same temporal information have high similarity and calculate the time similarity of entities according to their temporal information. If two entities are both the unique nearest neighbor to each other, then the two entities are regarded as alignment seeds. Experimental results show that the proposed unsupervised method outperforms the previous SOTA models on two datasets and competes with our supervised method STEA.

The main contributions of this paper are summarized as follows:

(1) We propose a simple GNN model combined with a temporal information matching mechanism, achieving significant performance with fewer parameters and less time.

(2) By assuming that entities with the same temporal information have high similarity, we propose a simple strategy to generate alignment seeds by selecting the nearest neighbor.

(3) Extensive experiments on public datasets indicate that our supervised method outperforms all SOTA methods. The performance of using automatically generated alignment seeds is close to that of the supervised method.

\section{Related Work}
\subsection{Translation-based Model}
Translation-based models learn the embeddings of entities by the triples of KGs, which interpret relation as translation operations from its head to its tail, such as \bm{$h+r \approx t$}. MTransE \cite{DBLP:conf/ijcai/ChenTYZ17} is the first model to use TransE \cite{DBLP:conf/nips/BordesUGWY13} for entity alignment which maps two KGs into different vector spaces. Entities with similar positions in two vector spaces are alignment pairs. In addition to using relational triples, JAPE \cite{DBLP:conf/semweb/SunHL17} also utilizes the attribute features of entities to learn the representation of entities to improve the performance of EA. BootEA \cite{DBLP:conf/ijcai/SunHZQ18} proposes a bootstrapping process by adding label likely alignment entities into training data iteratively to promote the EA results. MultiKE \cite{DBLP:conf/semweb/HuZSH19} represents entities based on muti-aspects of KGs to enhance the alignment. These methods use the triples to align entities independently and lack utilization of the global structural information of KGs.

\subsection{GNN-based Model}
Due to the powerful ability of GNN to model the structure of KGs, many GNN-based models have been proposed and have achieved good performance on the task of EA. GCN-Align \cite{DBLP:conf/emnlp/WangLLZ18} is the first method to use GCN \cite{DBLP:conf/iclr/KipfW17} for entity alignment. GCN-Align combines the attribute features of entities with the structural features of the KG and uses GCN to map the entities to a low-dimensional vector space so that the equivalent entities are close to each other in the space. However, GCN-Align does not effectively utilize the relational features in the knowledge graph. MRAEA \cite{DBLP:conf/wsdm/MaoWXLW20} proposes a meta relation aware EA method which leverages meta relation-aware embedding and relation-aware self-attention to align the KGs. RREA \cite{DBLP:conf/cikm/MaoWXWL20} designs a relational reflection transformation operation to preserve the similarity distributions of entities and integrate them into GNNs to facilitate EA. TEA-GNN \cite{DBLP:conf/emnlp/XuS021} adopts a similar idea from RREA and utilizes additional temporal information for EA between TKGs. It learns the embeddings of entities by a time-aware attention mechanism to promote EA. TREA \cite{DBLP:conf/www/XuSX022} uses a temporal relational attention mechanism to integrate relational and temporal features of links between nodes to enhance EA. In TKGs, the temporal information representing the same timestamps is uniform. We propose a simple GNN model combined with a temporal information matching mechanism, which is effective and efficient.

\subsection{Unsupervised Methods for Alignment Seeds Generation}
In recent years, some unsupervised methods have emerged to generate alignment seeds to solve the resource-consuming problem. IMUSE \cite{DBLP:conf/dasfaa/HeLQ0LZ0ZC19} uses both attribute triples and relation triples of KGs to collect seed with high text similarity. EVA \cite{DBLP:conf/aaai/0001CRC21} leverages the additional visual information of entities to create alignment seeds. PRASE \cite{DBLP:conf/ijcai/QiZCCXZZ21} uses PARIS to obtain the alignment seeds by literal attributes of the entities. These approaches incorporate additional information such as visual and text information of entities to obtain the alignment seeds. Inspired by these methods, we propose a simple alignment seeds generation method by utilizing the temporal information of entities.

\section{Problem Formulation}
TKGs store the real-world knowledge in the form of quadruples $(h,r,t,\tau)$.
A TKG is represented as $G=(E,R,T,Q)$, where $E$, $R$ and $T$ represent the sets of entities, relations and timestamps respectively, $Q\subset E\times R\times E\times T$ denotes the set of quadruples. Defining $G_1=(E_1,R_1,T_1,Q_1)$ and $G_2=(E_2,R_2,T_2,Q_2)$ to be two TKGs, $S=\{(e_{1_i},e_{2_j})|e_{1_i}\in E_1,e_{2_j}\in E_2\}$ is the set of alignment seeds between $G_1$ and $G_2$. Specifically, the timestamps in the two time set has been merged in a uniform time set $T^* = T_1 \cup T_2$. Therefore, the two TKGs can be renewed as $G_1=(E_1,R_1,T^*,Q_1)$ and $G_2=(E_2,R_2,T^*,Q_2)$ sharing the same set of timestamps. EA task aims to find new alignment entities set $P$ based on the alignment seeds $S$ between the two TKGs.

\section{The Proposed Approach}
\begin{figure*}
  \centering  
  \includegraphics[width=0.99\linewidth]{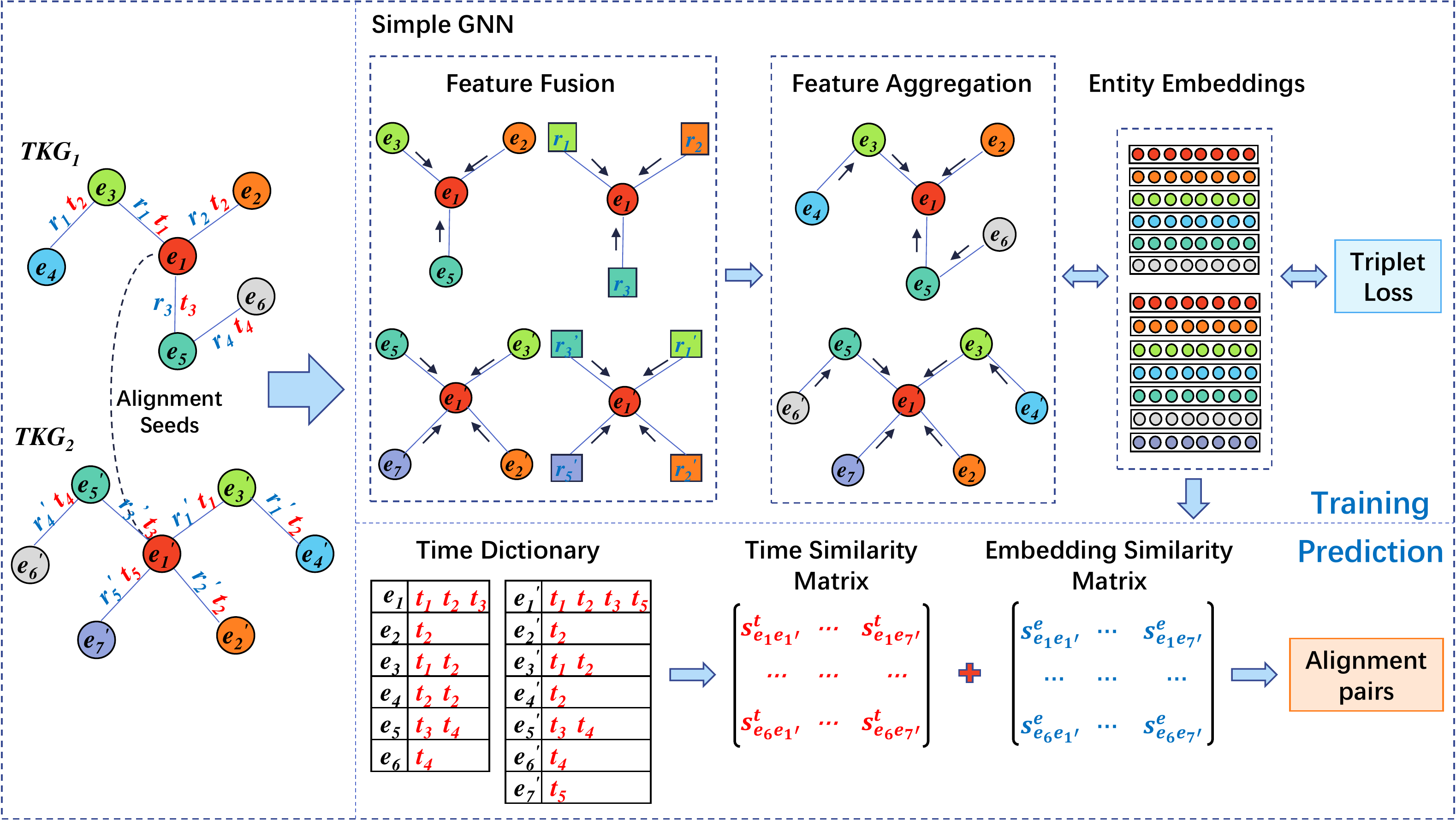}\\
  \caption{Framework of STEA. The solid line between entities represents the existent relations in TKGs, the dashed line represents the alignment seeds.}
  \label{fig:model}
  \vspace{-1em}
\end{figure*}

This research proposes a simple GNN model combined with a temporal information matching mechanism for EA between TKGs using the entity's structural information, relational information, and temporal information. During the training phase, the simple GNN is used to learn entities' structural and relational embeddings. In the prediction stage, the model predicts the new alignment entities by combining the embedding similarity and temporal matching degree of entities. Experimental results show that our model achieves good performance by balancing the embedding similarity with the time similarity of entities. Figure~\ref{fig:model} shows the framework of STEA.

\subsection{Simple Graph Neural Network}
The simple GNN model learns the embeddings of entities by using their structural and relational information. The model first fuses entities' structural features and relational features, then aggregates them through a simple graph convolution operation. Finally, it combines entity embeddings of different layers to obtain a global entity embedding. The sub-modules of the model include feature fusion, feature aggregation, and global embedding generation.

\subsubsection*{Feature Fusion}
Some studies \cite{DBLP:conf/emnlp/YangZSLLS19,DBLP:conf/cikm/MaoWXWL20} believe that both the structural information and the relational information are beneficial to the representation of entities. Therefore, we fuse the relational features and structural features to get semantic embeddings of entities. 
The structural feature and relational feature of entity $e_i$ are calculate by the following equations:
\begin{equation}
    \bm h^e_{e_i} = \frac { 1 } { | \mathcal{N}^e_{e_i} | } \sum_{ e_j \in \mathcal{N}^e_{e_i}} \bm h_{e_j}
    \label{eq1}
\end{equation}
\begin{equation}
    \bm h^r_{e_i} = \frac { 1 } { | \mathcal{N}^r_{e_i} | } \sum_{ r_j \in \mathcal{N}^r_{e_i}} \bm h_{r_j}
    \label{eq2}
\end{equation}
where $\mathcal{N}^e_{e_i}$ represents the neighboring entity set of $e_i$, $\bm h_{e_i}$ is the randomly generated initialization embedding of $e_i$, $\mathcal{N}^r_{e_i}$ represents the set of the relations around entity $e_i$, $\bm h_{r_i}$ is the randomly generated initialization embedding of $r_i$.

The fused feature of entity $e_i$ is obtained as follows:
\begin{equation}
  \bm h^{(1)}_{e_i} =[ \bm h^e_{e_i} || \bm h^r_{e_i} ]
  \label{eq3}
\end{equation}
where $||$ represents the concatenate operation, $\bm h^{(1)}_{e_i}$ represents the fused embedding of $e_i$ in layer $1$.
\subsubsection*{Feature Aggregation}
Each entity aggregates the fused features from its neighbors by a graph convolution operation. The aggregation can be performed efficiently by matrix operations. Let $\bm H^{(l)}$ represents the features of entities in layer $l$, $\bm A$ is the adjacency matrix with self-loops of TKGs, $\bm D$ is the diagonal node degree matrix, $\bm H^{(l+1)}$ is the embeddings of entities in layer $l+1$, The matrix operation of feature aggregation to get $\bm H^{(l+1)}$ is shown as follows:
\begin{equation}
  \bm H^{(l+1)}=\sigma(\bm D^{-1}\bm A \bm H^{(l)})
  \label{eq4}
\end{equation}
where $\sigma(\cdot)$ is a non-linear activation function like the ReLU.
\subsubsection*{Global-aware Embedding}
The GNN can capture multi-hop neighborhood information. We concatenate the entity embedding from different layers to get the global-aware embedding. The final embedding of entity $e_i$ is obtained as follows:
\begin{equation}
  \bm h^{mul}_{e_i} = [ \bm h^{(1)}_{e_i} ||\bm h^{(2)}_{e_i} ||...|| \bm h^{(L)}_{e_i}]
  \label{eq5}
\end{equation}
where $\bm h^{(1)}_{e_i}$ is the embedding of entity $e_i$ in layer $1$, $\bm h^{(L)}_{e_i}$ is the embedding of entity $e_i$ in layer $L$, $||$ represents the concatenate operation.

\subsection{Entity Alignment}
The model first learns the embeddings of entities by training with the alignment seeds, then calculates the embedding similarity matrix of the entities and combines the time similarity matrix to predict the new alignment entities. Iterative learning is performed by adding new alignment entities to the training set to find more alignment entities. 
\subsubsection*{Training}
We use the following triplet loss function to train the model:
\begin{small}
\begin{equation}
L = \sum_{(e_i,e_j)\in P}max( dist(\bm e_i,\bm e_j)-dist(\bm e'_i,\bm e'_j)+\lambda,\;0)
\label{eq6}
\end{equation}
\end{small}
where $(e_i,e_j)$ represent the positive pair, $(e'_i,e'_j)$ represent the negative pair by randomly replacing one of $(e_i,e_j)$. $\lambda$ represents the margin hyper-parameter, $dist(\cdot)$ represent the Manhattan distance:
\begin{equation}
  dist(\bm e_i,\bm e_j) = \|\bm h^{mul}_{\bm e_i} - \bm h^{mul}_{\bm e_j}\|_1
  \label{eq7}
\end{equation}
\subsubsection*{Time Dictionary Construction}
There are two types of temporal information in the datasets, one is the time point $\tau$ and another is time interval $[\tau_s,\tau_e]$, where $\tau_s$ denote the start time and $\tau_e$ denote the end time. We construct the time dictionary $Dic$ of the entities by using the quadruples. If the quadruple is $(h,r,t,\tau)$, then append $\tau$ to the dictionary $Dic_h$ of $h$ and dictionary $Dic_t$ of $t$. If the quadruple is $(h,r,t,[\tau_s,\tau_e])$, then append $\tau_s$ and $\tau_e$ to the dictionary $Dic_h$ and $Dic_t$ . 
\subsubsection*{Temporal Information Matching Mechanism}
We use the time dictionary mentioned above to obtain the time similarity matrix by calculating the matching degree of temporal information in the time dictionary of each entity. Specifically, for two entities $e_i \in G_1$ and $e_j \in G_2$, the time dictionaries of them are $Dic_{e_i} = [\tau^1_{e_i}, \tau^2_{e_i}, ..., \tau^m_{e_i}]$ and $Dic_{e_j} = [\tau^1_{e_j}, \tau^2_{e_j}, ..., \tau^n_{e_j}]$, where $m$ and $n$ represent the numbers of temporal information in $Dic_{e_i}$ and $Dic_{e_j}$ respectively. Let $c$ denote the number of same items in $Dic_{e_i}$ and $Dic_{e_j}$, the time similarity $s^t_{e_ie_j}$ of $e_i$ and $e_j$ is 
% equal to $ (c \times 2) / (m + n )$. 
calculated as follows:
\begin{equation}
    s^t_{e_ie_j} = \frac { c \times 2 } { m + n } 
    \label{eq8}
\end{equation}

For each entity in the two TKGs, by calculating the matching degree of temporal information from each other, we get the time similarity matrix of entities in different TKGs.
\subsubsection*{Prediction}
We adopt a similar method with RREA \cite{DBLP:conf/cikm/MaoWXWL20} which uses the Cross-domain Similarity Local Scaling (CSLS) \cite{DBLP:conf/iclr/LampleCRDJ18} to predict the alignment entities. Since TKGs have additional temporal information, which is used to calculate the time similarity of entities, we predict alignment entities by combining the embedding similarity and time similarity of entities. The final similarity of entities can be expressed as follows:
\begin{equation}
s_{e_ie_j}=(1-\alpha) \times s^e_{e_ie_j}+\ \alpha\times s^t_{e_ie_j}
  \label{eq9}
\end{equation}
where $s^e_{e_ie_j}$ is the embedding similarity of $e_i$ and $e_j$, it can be simply calculated by the inner product of the embeddings of $e_i$ and $e_j$. $s^t_{e_ie_j}$ is the time similarity of $e_i$ and $e_j$, $\alpha$ is a balance hyper-parameter. 
\subsubsection*{Iterating Learning}
We adopt the bi-directional iterative strategy proposed by MRAEA \cite{DBLP:conf/wsdm/MaoWXLW20}. The difference from this method is that we also combine the temporal information to calculate the similarity between entities in addition to using entity embedding. If the entities $e_i$ and $e_j$ are mutually nearest neighbors, then the pair $(e_i, e_j)$ is considered new alignment entities and will be added into the training set of the next iteration.
Since an iteration contains many epochs of training (1200 in this paper), the pseudo-supervised seeds generated after one iteration help improve the model's performance.

\subsection{Unsupervised Alignment Seeds Generation}
In the real world, alignment seeds are often difficult to obtain due to the high cost of manual annotations. Assuming that the entities with the same temporal information in the two TKGs are the most similar, we propose a strategy to generate the alignment seeds using the temporal information of entities. The generated alignment seeds $(e_i,e_j)$ should satisfy the following two criteria: (1) For $e_i$, there is and only one $e_j$ that exactly matches the temporal information of $e_i$. (2) They are mutually and uniquely nearest neighbors of each other. The detailed alignment seeds generation process is described in Algorithm~\ref{alg:algorithm}.
\begin{algorithm}[tb]
\caption{Algorithm for alignment seeds generation}
\label{alg:algorithm}
\textbf{Input}: time similarity matrix of entities $\bm S^t$.

\textbf{Output}: Alignment seeds set $AS$.

\begin{algorithmic}[1] %[1] enables line numbers
\STATE Let $AS=\{ \ \}, index=\{ \ \}$.
\FOR{$i \in \bm S^t[i]$}
\IF {$(len(argmax(\bm S^t[i]) == 1)\qquad\qquad\ \ $ $ \qquad\qquad\qquad \ \& \ (max(\bm S^t[i])==1) $}
\STATE $index \gets (i,argmax(\bm S^t[i])) \cup index$
\ENDIF
\ENDFOR
\FOR{$(a,b) \in index$}
\IF {$(\bm S^t[a][b]\ ==\ \bm S^t[b][a])$}
\STATE $AS \gets (a,b) \cup AS$
\ENDIF
\ENDFOR
\STATE \textbf{return} $AS$
\end{algorithmic}
\end{algorithm}

\section{Experiments}
\subsection{Data sets}
We experiment on four datasets for entity alignment between temporal knowledge graphs in English.
\begin{table*}
\centering
\resizebox{\textwidth}{!}{
\begin{tabular}{c c c c c c c c c c}
\hline
{Dataset} & {$|E_1|$} & {$|E_2|$} & {$|R_1|$} & {$|R_2|$} & {$|T|$} & {$|Q_1|$} & {$|Q_2|$} & {$|P|$} & {$|S|$}\\
\hline
\textbf{DICEWS-1K} & {9,517} & {9,537} & {247} & {246} & {4,017} & {307,552} & {307,553} & {7,566} & {1,000}\\
\textbf{DICEWS-200} & {9,517} & {9,537} & {247} & {246} & {4,017} & {307,552} & {307,553} & {8,366} & {200}\\
\textbf{YAGO-WIKI50K-5K} & {49,629} & {49,222} & {11} & {30} & {245} & {221,050} & {317,814} & {44,172} & {5,000}\\
\textbf{YAGO-WIKI50K-1K} & {49,629} & {49,222} & {11} & {30} & {245} & {221,050} & {317,814} & {48,172} & {1,000}\\
\hline
\end{tabular}}
\caption{Statistics of YOGO-WIKI50K and DICEWS. $|\cdot|$ represents the numbers.}
\label{tab:datasets}
\end{table*}

(1) \textbf{DICEWS} \cite{DBLP:conf/emnlp/XuS021}: This dataset is built from ICEWS05-15 \cite{DBLP:conf/emnlp/Garcia-DuranDN18} which contains events during 2005 to 2015. There are \textbf{DICEWS-1K} with 1K alignment seeds and \textbf{DICEWS-200} with 200 in it.

(2) \textbf{YAGO-WIKI50K} \cite{DBLP:conf/emnlp/XuS021}: This dataset extracts the top 50,000 entities according to their frequencies in Wikidata from tkbc \cite{DBLP:conf/iclr/LacroixOU20} which has temporal information, and then link the Wikidata entities to their equivalent YAGO entities by the mappings of Wikidata QIDs to YAGO instances \footnote{http://resources.mpi-inf.mpg.de/yago-naga/yago3.1/}. There are two subsets \textbf{YAGO-WIKI50K-5K} and \textbf{YAGO-WIKI50K-1K} in the datasets. The only difference between the subsets is that one has 5K alignment seeds, and the other has 1K.
 
The statistics of these datasets are listed in Table~\ref{tab:datasets}.

\subsection{Baselines}
We compare our method with the following three groups of methods:

(1) {\bf Translation-based Model}: MTransE \cite{DBLP:conf/ijcai/ChenTYZ17}, JAPE \cite{DBLP:conf/semweb/SunHL17}, BootEA \cite{DBLP:conf/ijcai/SunHZQ18}. These methods are strong baseline based on TransE.

(2) {\bf GNN-based Model}: GCN-Align \cite{DBLP:conf/emnlp/WangLLZ18}, MRAEA \cite{DBLP:conf/wsdm/MaoWXLW20}, RREA \cite{DBLP:conf/cikm/MaoWXWL20}. These are SOTA GNN-based models for EA between KGs without temporal information. 

(3) {\bf Time-aware Model}: TEA-GNN \cite{DBLP:conf/emnlp/XuS021}, TREA \cite{DBLP:conf/www/XuSX022}. These are SOTA GNN-based models for EA between TKGs. 

The main results of these models reported in the paper are from TREA \cite{DBLP:conf/www/XuSX022} except BootEA \cite{DBLP:conf/ijcai/SunHZQ18} and RREA \cite{DBLP:conf/cikm/MaoWXWL20}, because the source code of the two models uses an iterative strategy, while TREA uses a non-iterative version. The experiments of these two models are implemented based on their publicly available resource codes. All experiments are conducted on a workstation with a GeForce RTX 3090 GPU and an AMD EPYC 7502 32-Core Processor CPU and 128GB memory. The code and datasets are available online  \footnote{https://github.com/lcai2/STEA}.

\begin{table*}
\centering
\resizebox{\textwidth}{!}{
\begin{tabular}{l c c c c c c c c c c c c c}
\hline
\multirow{2}{*}{Models} & \multicolumn{3}{c}{\bf DICEWS-1K} & \multicolumn{3}{c}{\bf DICEWS-200} & \multicolumn{3}{c}{\bf YAGO-WIKI50K-5K} &  \multicolumn{3}{c}{\bf YAGO-WIKI50K-1K}\\
\cmidrule(rl){2-4}%绘制第2列和第4列的横线，留空
\cmidrule(rl){5-7}
\cmidrule(rl){8-10}
\cmidrule(rl){11-13}
& {MRR} & {Hits@1} &  {Hits@10} & {MRR} & {Hits@1} &  {Hits@10} & {MRR} & {Hits@1} &  {Hits@10} & {MRR} & {Hits@1} &  {Hits@10}\\
\hline
{MTransE} & {.150} & {.101} & {.241} & {.104} & {.067} & {.175} & {.332} & {.242} & {.477} & {.033} & {.012} & {.067}\\
{JAPE} & {.198} & {.144} & {.298} & {.138} & {.098} & {.210} & {.345} & {.271} & {.488} & {.157} & {.101} & {.262}\\
{AlignE} & {.593} & {.508} & {.751} & {.303} & {.222} & {.457} & {.800} & {.756} & {.883} & {.618} & {.565} & {.714} \\
{BootEA} & {.670} & {.598} & {.796} & {.614} & {.546} & {.737} & {-} & {-} & {-} & {-} & {-} & {-} \\
\hline
{GCN-Align} & {.291} & {.204} & {.466} & {.231} & {.165} & {.363} & {.581} & {.512} & {.711} & {.279} & {.217} & {.398}\\
{MRAEA} & {.745} & {.675} & {.870} & {.564} & {.476} & {.733} & {.848} & {.806} & {.913} & {.685} & {.623} & {.801}\\
{RREA} & {.840} & {.795} & {.917} & {.823} & {.773} & {.911} & {.913} & {.887} & {.955} & {.870} & {.836} & {.929}\\
\hline
{TEA-GNN} & {.911} & {.887} & {.947} & {.902} & {.876} & {.941} & {.909} & {.879} & {.961} & {.775} & {.723} & {.871}\\
{TREA} & {.933} & {.914} & {\underline{.966}} & {.927} & {.910} & {.960} & {\underline{.958}} & {\underline{.940}} & {\underline{.989}} & {.885} & {.840} & {.937}\\
\hline
{STEA*} & {\underline{.941}} & {\underline{.928}} & {.960} & {\underline{.941}} & {\underline{.927}} & {\underline{.961}}  & {.954} & {.935} & {.986} & {\underline{.916}} & {\underline{.887}} & {\underline{.966}}\\
{STEA} & {\bf.954} & {\bf.945} & {\bf.967} & {\bf.954} & {\bf.943} & {\bf.968} & {\bf.974} & {\bf.961} & {\bf.992} & {\bf.962} & {\bf.943} & {\bf.989}\\
\hline
\end{tabular}}
\caption{ Experimental results of EA on DICEWS and YOGO-WIKI50K. AlignE is the non-iterative version of BootEA. - means the results are not obtained. STEA* represents the mothed without iteration. The best results are written in bold. Underline indicate the sub-optimal results. }
\label{tab:results}
\end{table*}

\subsection{Experimental Setup}
For a fair comparison, we use the fixed training set and validation set provided by TEA-GNN \cite{DBLP:conf/emnlp/XuS021}. TREA \cite{DBLP:conf/www/XuSX022} uses the same datasets, but the source code is not yet provided. We also provide a non-iterative model STEA* to compare with the SOTA models.

Following convention, Hits@$k$ $(k=1,10)$ and mean reciprocal rank (MRR) are used as evaluation metrics. Hits@$k$ reports the proportion of correct alignment pairs to the top $k$ potential entities. In particular, Hits@1 represents accuracy. MRR is calculated as the average of the reciprocal ranks of the results. The higher the Hits@$k$ and MRR, the better the performance.

For all datasets, we use a same default setting: the dimensionality for embedding $d=100$; depth of GNN layers $L = 2$; margin $\gamma = 3$; dropout\_rate $ dr = 0.3$; balance factor $\alpha = 0.3$; iterations $k = 5$; The number of epochs is $1200$ and RMSprop is adopted to optimize the model with learning rate set to 0.005. The reported performance is the average of five independent training runs.
\subsection{Results and Analysis}
\subsubsection*{Main Results}
Table~\ref{tab:results} show the main results of the experiment.
It can be seen that STEA with iteration achieves the best results on all datasets compared to all methods on all the evaluation metrics. Especially compared with the SOTA method TREA for EA between TKGs, STEA exceeds by at least 2.2\% on Hits@1. The improvement scores of Hits@1 are 3.4\%, 3.6\%, 2.2\%, 12.3\%, respectively. Without iteration, STEA* still outperforms TREA on most datasets and metrics, and the gaps in other experimental results are very tiny, at most no more than 0.06\%.
It indicates the effectiveness of combining the embedding similarity and time similarity to predict alignment entities. STEA achieves the best results and proves that iterative strategies could improve performance. Our method achieves remarkable performance using the simple GNN model combined with a temporal information matching mechanism.

\subsubsection*{Ablation Study}
The STEA consists of three key components: 
(1) Relational Feature Fusion (RFF); 
(2) Global-Aware Representation (GAR); 
(3) Time Similarity Matrix (TSM). 
We remove these components from STEA individually to demonstrate their effectiveness.

Table~\ref{tab:ablation} show the ablation study of the experiment. It can be seen from the table that the performance of STEA is degraded after removing a module. Each module contributes differently to the two datasets due to their different characteristics. Without relational feature fusion, EA performance drops slightly on YAGO-WIKI50K (STEA vs. STEA-RFF) because there are few relations in the datasets and decreases more on DICEWS with more relations. This shows the importance of relational features in heterogeneous TKGs with multiple relationships. STEA-GAR Remove concatenating the representations of different layers, Hits@1 declines by about 3\% on DICEWS and 5\% on YAGO-WIKI50K, validating the effectiveness of gathering the embedding of multi-hop neighborhood. Compared with STEA, without using the temporal information, STEA-TSM does hurt STEA in all metrics, indicating the importance of using the temporal matching mechanism. The performance drops more on DICEW than YAGO-WIKI50K, suggesting that the former is more relevant to temporal information.
\begin{table*}
\centering
\resizebox{\textwidth}{!}{
\begin{tabular}{l c c c c c c c c c c c c}
\hline
\multirow{2}{*}{Models} & \multicolumn{3}{c}{\bf DICEWS-1K} & \multicolumn{3}{c}{\bf DICEWS-200} & \multicolumn{3}{c}{\bf YAGO-WIKI50K-5K} &  \multicolumn{3}{c}{\bf YAGO-WIKI50K-1K}\\
\cmidrule(rl){2-4}%绘制第2列和第4列的横线，留空
\cmidrule(rl){5-7}
\cmidrule(rl){8-10}
\cmidrule(rl){11-13}
& {MRR} & {Hits@1} &  {Hits@10} & {MRR} & {Hits@1} &  {Hits@10} & {MRR} & {Hits@1} &  {Hits@10} & {MRR} & {Hits@1} &  {Hits@10}\\
\hline
{STEA} & {\bf.954} & {\bf.945} & {\bf.967} & {\bf.954} & {\bf.943} & {\bf.968} & {\bf.974} & {\bf.961} & {\bf.992} & {\bf.962} & {\bf.943} & {\bf.989}\\
{STEA-RFF} & {.923} & {.910} & {.944} & {.924} & {.910} & {.946}  & {.967} & {.953} & {.987} & {.948} & {.925} & {.981}\\
{STEA-GAR} & {.932} & {.916} & {.952} & {.934} & {.918} & {.956} & {.945} & {.918} & {.984} & {.927} & {.890} & {.977}\\
{STEA-TSM} & {.792} & {.741} & {.888} & {.742} & {.674} & {.863} & {.932} & {.910} & {.967} & {.881} & {.843} & {.939}\\
\hline
\end{tabular}}
\caption{Ablation study of STEA on all datasets}
\label{tab:ablation}
\end{table*}

\subsubsection*{Complexity analysis}
\begin{figure}
  \centering  
  \includegraphics[scale=0.53]{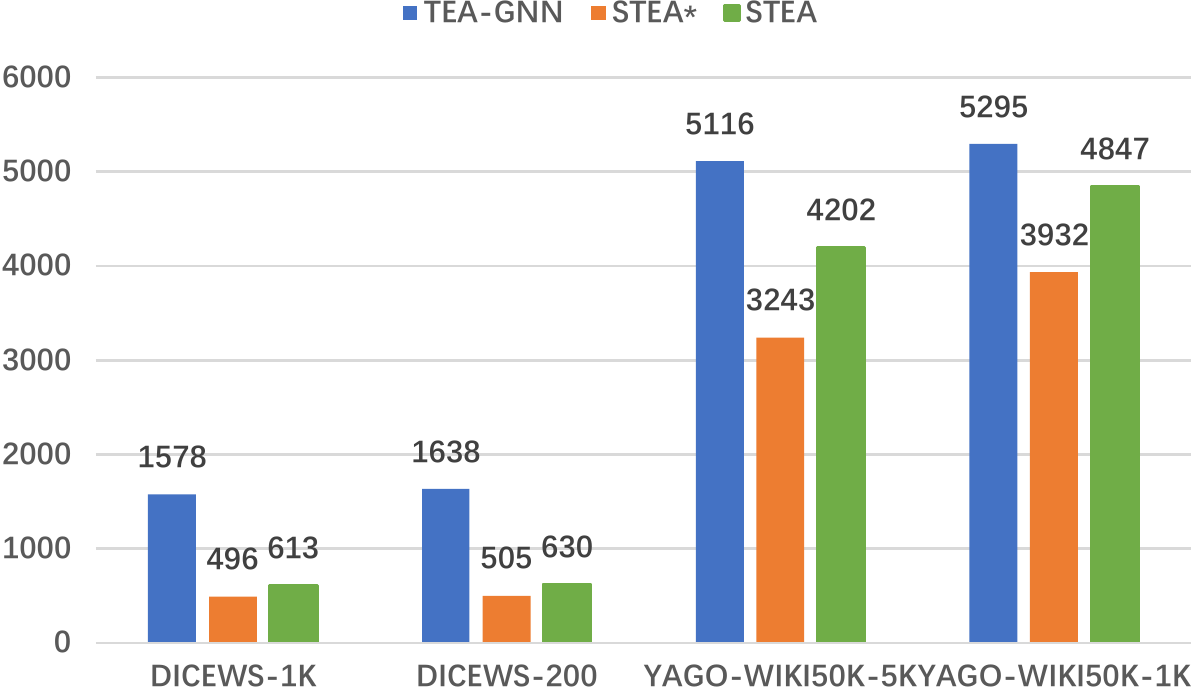}\\
  \caption{ Time costs of methods (seconds).}
  \label{fig:time_costs}
  \vspace{-1em}
\end{figure}

Figure\ref{fig:time_costs} reports the overall time costs of our methods with TEA-GNN \cite{DBLP:conf/emnlp/XuS021} on each dataset, including data loading, pre-processing, training, and evaluating. The results are obtained by directly running the source code provided by the author. 
As shown in Figure~\ref{fig:time_costs}, the time cost of STEA* is the lowest. The time cost of STEA is higher than STEA* since using the iteration, but still less than TEA-GNN, which indicates the efficiency of our model.

The total number of parameters in the training phase of our model is equal to $(|E| + |R|) \times d$, which is less than the existing SOTA model. 
The number of parameters of each model compared in our experiments is shown in Table ~\ref{tab:complexity}, 
\begin{table}
\centering
\resizebox{\linewidth}{!}{
\begin{tabular}{l l}
\hline
{Model} & {Number of Parameters}\\
\hline
{MTransE} &  {$O((|E| + |R|) \times d)$} \\
{JAPE} & {$O((|E| + |R|) \times d)$} \\
{BootEA} & {$O((|E| + |R|) \times d)$}\\
{GCN-Align} & {$O(|E| \times d + d \times d \times L)$}\\
{MRAEA} &  {$O((|E| + |R|\times 2) \times d)$}\\
{RREA} & {$O((|E| + |R|\times 2) \times d + 3 \times d \times L)$}\\
{TEA-GNN} & {$O((|E| + |R|\times 2 + |T|) \times d + 3 \times d \times L \times 2)$}\\
{TREA} & {$O((|E| + |R|\times 2 + |T|) \times d + 4 \times d \times L \times 2)$}\\
{STEA} & {$O((|E| + |R|) \times d)$}\\
\hline
\end{tabular}}
\caption{The total number of parameters in the training phase of the models compared in the paper.}
\label{tab:complexity}
\end{table}
where $|E|$ represents the total entities of the two TKGs, $|R|$ represents the total relations of the two TKGs, $d$ represents the embedding dimension of the parameters, $L$ represents the layers of GNN, $|T|$ represents the total temporal information of the two TKGs.

\subsubsection*{Hyper-parameter Analysis}
\begin{figure}
  \centering  
  \includegraphics[scale=0.56]{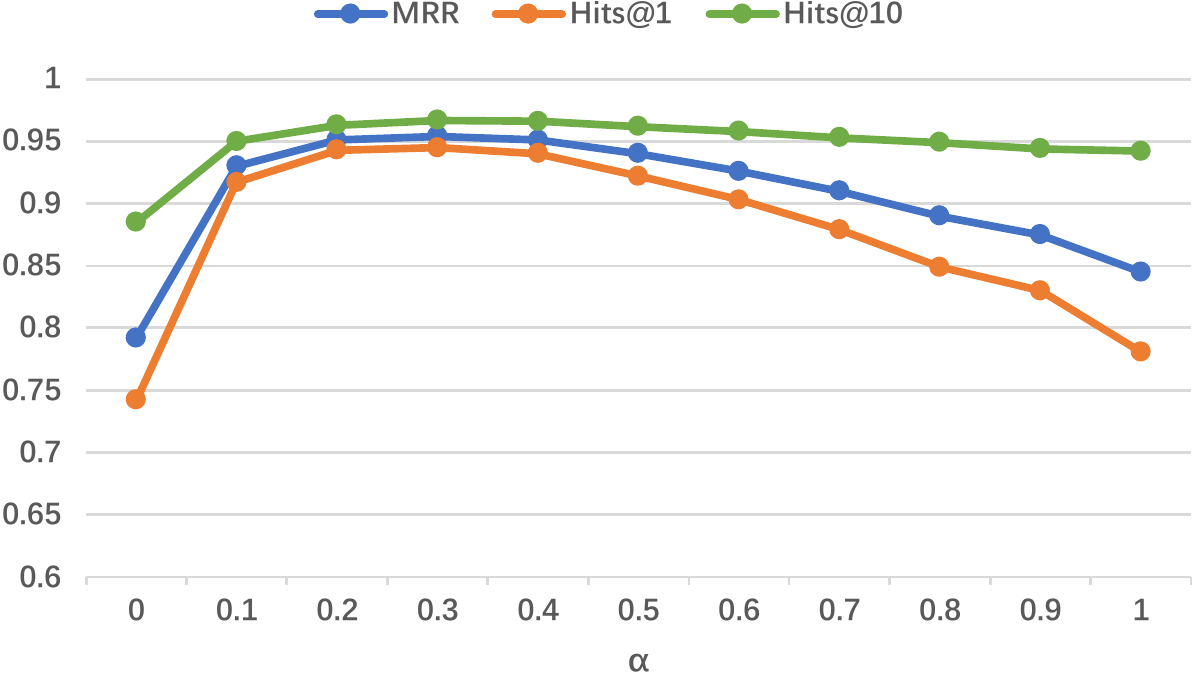}\\
  \caption{ Experimental results with different balance factor $\alpha$ on DICEWS-1K.}
  \label{fig:alpha}
  \vspace{-1em}
\end{figure}
In order to investigate the effects of hyper-parameters on the performance of STEA, we conduct the following experiments on DICEWS-1K: (1) The performance with different balance factors $\alpha$. (2) The entity alignment results in different layers $l$. (3) The Hits@1 values with different embedding dimensions $d$.

The balance factor $\alpha$ is used to balance the weight of the entity's embedding similarity matrix and the time similarity matrix in our approach. We set it in range $0\sim1$ with interval $0.1$. As shown in Figure~\ref{fig:alpha}, the performance of EA only using the embedded similarity matrix ($\alpha=0$) is lower than that of only using the time similarity matrix ($\alpha=1$). It shows that the temporal information in the dataset is highly matched. The model's performance is optimal when $\alpha=0.3$ and similar in the range of $0.1\sim0.5$. 

For more analysis about hyper-parameters layers $l$ and dimension $d$, please refer to Appendix~\ref{sec:hyper-parameters}.

\subsubsection*{Unsupervised Experimental Results}
\begin{figure}
  \centering  
  \includegraphics[scale=0.56]{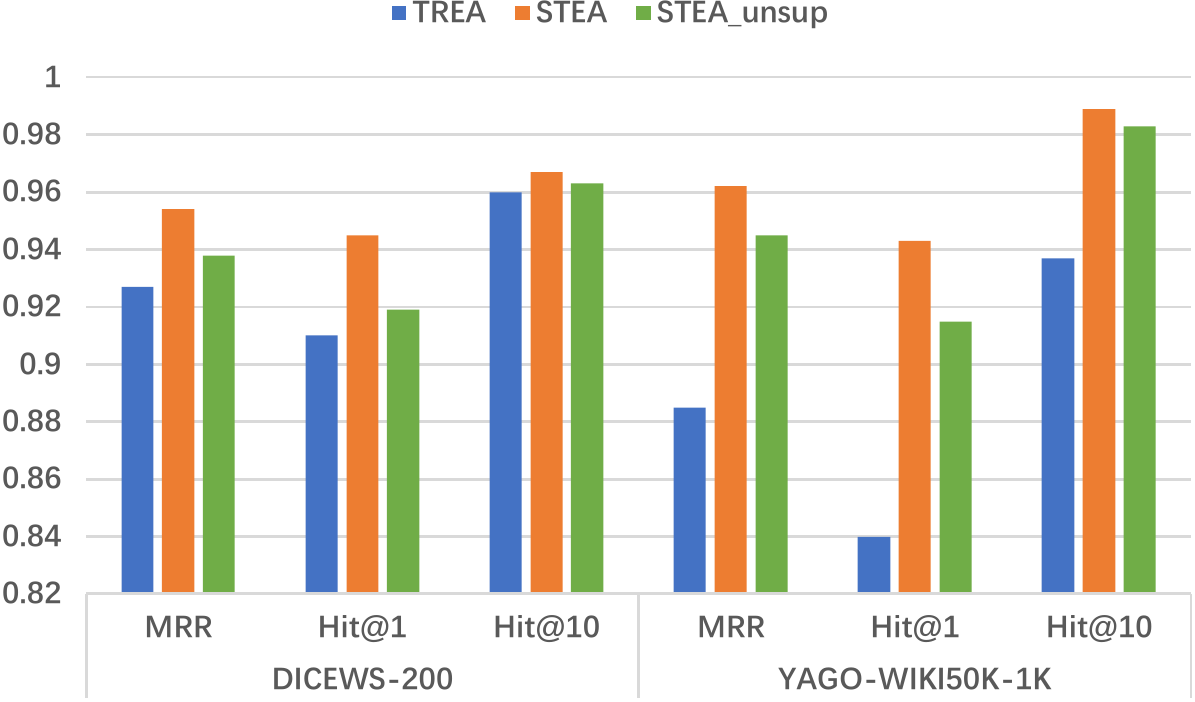}\\
  \caption{Experimental Results with unsupervised alignment seeds.}
  \label{fig:unsuper}
  \vspace{-1em}
\end{figure}

We adopt the approach mentioned in section 4.3 to obtain the alignment seeds and conduct experiments on DICEWS-200 and YAGO-WIKI50K-1K to compare against the supervised method TREA \cite{DBLP:conf/www/XuSX022} and STEA. The experimental results in Figue~\ref{fig:unsuper} show that our unsupervised method STEA$_{unsup}$ still outperforms TREA. The performance of STEA$_{unsup}$ on the two datasets is slightly lower than our supervised method STEA, indicating that only using the temporal matching degree of entities to obtain alignment seeds on the KGs may introduce noise.

\subsubsection*{Case Study}
In order to study the strengths and weaknesses of our approach compared to previous works, we illustrate some cases that STEA predicts from the test sets of YOGO-WIKI50K-1K and DICEWS-1K compared to TEA-GNN. Details are described in Appendix~\ref{sec:cases}.

\section{Conclusion}
This paper proposes a simple GNN model combined with a temporal information matching mechanism for EA between TKGs, which achieves significant performance with less time and fewer parameters.

By assuming that entities with the same temporal information have high similarity, we propose a method to generate the alignment seeds by temporal information. The unsupervised method outperforms the previous models on two datasets and is competitive with our supervised method.

The paper does not discuss the cases where the KGs might employ different schemas for representing time. We will further explore this in future work.

\section{Acknowledgements}
This work has been supported by Pudong New Area Science \& Technology Development Fund.
\bibliography{anthology,custom}

\appendix

\begin{table*}[htbp]
\centering
\resizebox{\textwidth}{!}{
\begin{tabular}{p{2.9cm} p{2cm} l}
\hline
{} & {Entities} & {Similar Quadruples Involving the Entities}\\
\hline
\multirow{4}{2.9cm}{Entity to be Aligned (in YAGO)} & \multirow{4}{2cm}{<Neymar>} 
& {\emph{<Neymar>,<playsFor>,<Brazil\_national\_under\_17\_football\_team>,(2009,inf)}}\\
& {} & {\emph{<Neymar>,<playsFor>,<Brazil\_national\_under\_20\_football\_team>,(2011,inf)}} \\
& {} & {\emph{<Neymar>,<playsFor>,<Brazil\_national\_under\_23\_football\_team>,(2012,inf)}} \\
& {} & {...} \\
\multirow{4}{2.9cm}{Prediction in TEA-GNN  (in WIKI)} & \multirow{4}{2cm}{Ademilson (Q2756361)} & {\emph{Ademilson(Q2756361),member of sports team,Brazil\_national\_under\_17\_football\_team(Q2402747),(2011,2011)}}\\
& {} & {\emph{Ademilson(Q2756361),member of sports team,   Brazil\_national\_under\_20\_football\_team(Q2308235),(2012,2013)}}\\
& {} & {\emph{Ademilson(Q2756361),member of sports team,   Brazil\_national\_under\_23\_football\_team(Q899189),(2014,inf)}}\\
& {} & {...} \\
\multirow{4}{2.9cm}{Prediction in STEA (in WIKI)} & \multirow{4}{2cm}{Neymar (Q142794)} 
& {\emph{Neymar (Q142794),member of sports team,Brazil\_national\_under\_17\_football\_team(Q2402747),(2009,2009)}}\\
& {} & {\emph{Neymar (Q142794),member of sports team,   Brazil\_national\_under\_20\_football\_team(Q2308235),(2011,2011)}} \\
& {} & {\emph{Neymar (Q142794),member of sports team,   Brazil\_national\_under\_23\_football\_team(Q899189),(2012,2012)}} \\
& {} & {...} \\
\hline
\end{tabular}}
\caption{The case of different predictions in STEA and TEA-GNN.}
\label{tab:goodcase}
\end{table*}

\begin{table*}[htbp]
\centering
\resizebox{\textwidth}{!}{
\begin{tabular}{p{3.5cm} p{3.0cm} l}
\hline
{} & {Entities} & {Similar Quadruples Involving the Entities}\\
\hline
\multirow{2}{3.5cm}{Entity to be Aligned (in ICEWS$_1$)} & 
\multirow{2}{3.0cm}{Bishop (India)} & 
\multirow{2}{*}{\emph{Head of Government (India), Make statement, Bishop (India), 462}}\\
{} & {} & {}\\

\multirow{2}{3.5cm}{Prediction in STEA \\(in ICEWS$_2$)} & 
\multirow{2}{3.0cm}{Electoral Alliance (India)} & 
\multirow{2}{*}{\emph{Head of Government (India),Make statement,Electoral Alliance (India),462}}\\
{} & {} & {}\\
\hline
\multirow{2}{3.5cm}{Entity to be Aligned (in YAGO)} &
\multirow{2}{3.0cm}{<William\_Travilla>} & \multirow{2}{*}{\emph{<William\_Travilla>,<isMarriedTo>,<Dona\_Drake>,(1944,1989)}}\\
{} & {} & {}\\

\multirow{2}{3.5cm}{Prediction in STEA \\(in WIKI)} & 
\multirow{2}{3.0cm}{Dona\_Drake (Q3035962)} & 
\multirow{2}{*}{\emph{Dona\_Drake(Q3035962), spouse, William\_Travilla(Q945402),(1944,1989)}}\\
{} & {} & {}\\
\hline
\end{tabular}}
\caption{The bad case of predictions in STEA.}
\label{tab:badcase}
\end{table*}

\section{Analysis of Hyper-parameters}
\label{sec:hyper-parameters}

\begin{figure}[htbp]
  \centering  
  \includegraphics[scale=0.6]{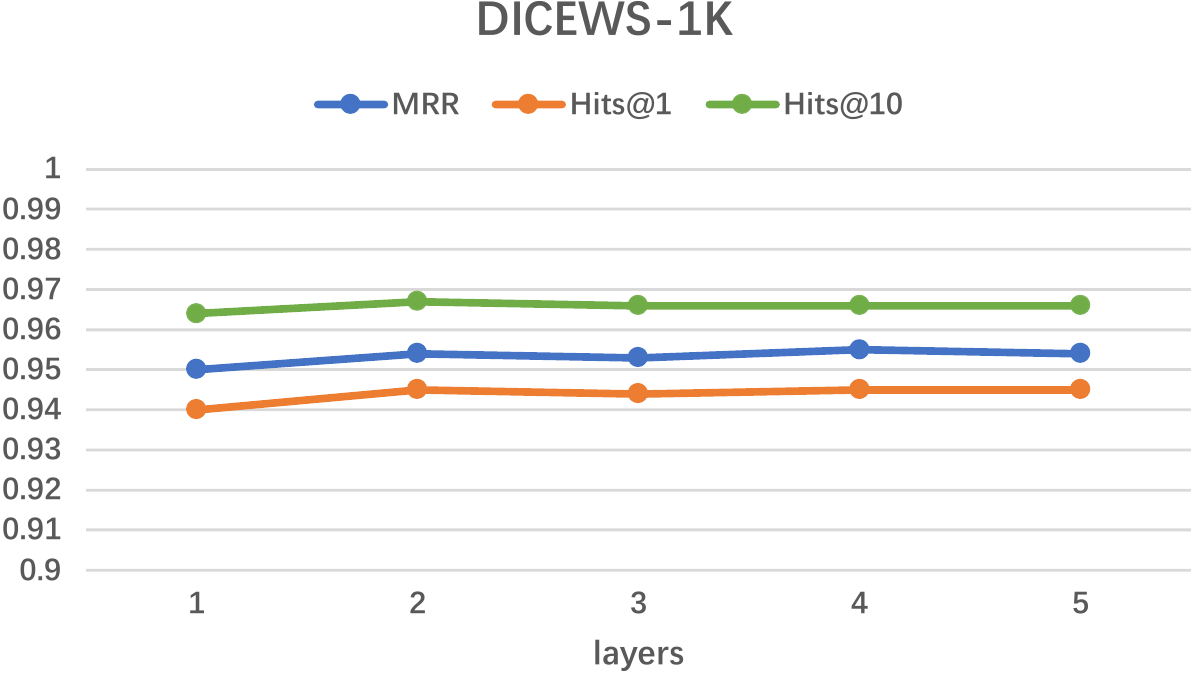}\\
  \caption{Entity alignment results with different layers of STEA on DICEWS-1K.}
  \label{fig:layer}
  \vspace{-1em}
\end{figure}

Figure~\ref{fig:layer} presents the entity alignment results with different layers of STEA on DICEWS-1K. It can be seen that the performance gap of the models with different layers is tiny, and STEA with 2 layers achieves the best performance. When stacking more layers, the performance does not improve. Increasing the layers only introduces more computation, not better performance. 

\begin{figure}[htbp]
  \centering  
  \includegraphics[scale=0.6]{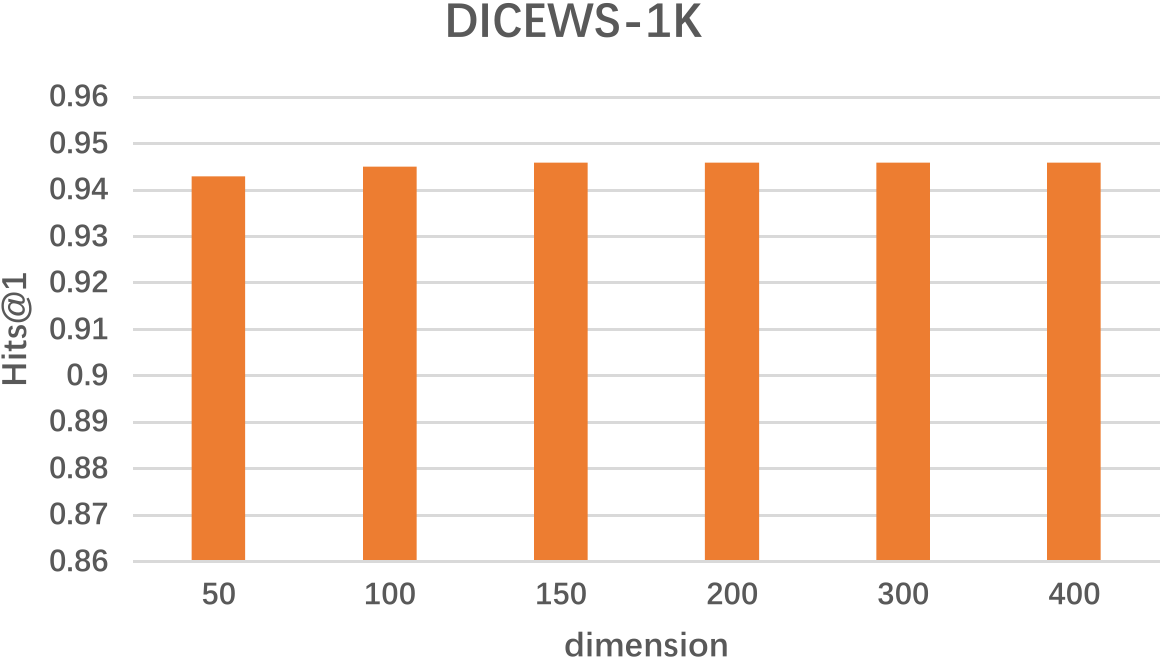}\\
  \caption{Hits@1 performances of different dimensions on DICEWS-1K.}
  \label{fig:dimension}
  \vspace{-1em}
\end{figure}

Figure~\ref{fig:dimension} reports the Hits@1 performances with embedding dimension $d$ from $50$ to $400$. 
The performance of STEA is not greatly affected by the dimension. As the dimension increases, the performance gain is small. Good performance is achieved even when dimension is $50$. We chose $100$ as the final dimension, which has high performance and small space.

\section{Case Study }
\label{sec:cases}

We illustrate some predictions of the STEA to analyze its advantages and problems. Table~\ref{tab:goodcase} shows an example that STEA gives a prediction different from TEA-GNN in YOGO-WIKI50K-1K. It can be seen from the case that TEA-GNN wrongly aligns \emph{Neymar} and \emph{Ademilson} from $G_1$ and $G_2$, because these two entities have similar connected links and temporal information in TKG1 and TKG2. Some links respective to these two entities have the same linked entities and relation types but different temporal information. The TEA-GNN wrongly identifies them as alignment entities by the time-aware attention mechanism. The temporal information matching mechanism of STEA can correctly distinguish these two entities since their timestamps are not equal.

Due to the temporal information matching mechanism, STEA will wrongly align entities with the same timestamps. There are some cases in Table~\ref{tab:badcase}. In the first case, entity \emph{Bishop (India)} from $G_1$ and entity \emph{Electoral Alliance (India)} from $G_2$ of DICEWS-1K are predicted as the alignment pairs because these two entities have the same links and temporal information in TKG1 and TKG2. STEA regards them as alignment entities due to their high similarity of structure and timestamps. In the second case, STEA mistakenly regards the entity \emph{<William\_Travilla>} in YAGO and the entity \emph{Dona\_Drake (Q3035962)} in WIKI as the same person since these two-person are a couple and get married on the same day. The model will misidentify these entities with symmetric relationships as alignment entities. TEA-GNN also has the same problem. It will be addressed in future work. 

\end{document}